\documentclass[]{aidata}

\usepackage[toc,page,header]{appendix}
\usepackage{minitoc}
\usepackage{cleveref} 
\usepackage{subcaption}
\usepackage{booktabs}
\usepackage{algorithm}
\usepackage{algpseudocode}

\usepackage{natbib}
\usepackage{latexsym}

\usepackage{url}
\usepackage{amssymb}
\usepackage[utf8]{inputenc}
\usepackage{microtype}
\usepackage{booktabs}
\usepackage{pifont} 
\usepackage{multirow}
\usepackage{makecell}
\usepackage{paralist}
\usepackage{xspace}
\usepackage{color}
\usepackage{xcolor}
\usepackage{colortbl}
\usepackage{adjustbox}
\usepackage{hyperref} 
\usepackage[edges]{forest}
\usepackage{tikz} 
\usepackage{caption}
\usepackage{amsfonts}

\hypersetup{
    colorlinks,
    linkcolor={blue!80!black},
    citecolor={blue!80!black},
}
\tikzset{
    root/.style =             {align=center, text width=1cm, rounded corners=3pt, line width=0.3mm, fill=gray!10, draw=gray!80, font=\small},
    demographic/.style =         {align=center, text width=1.8cm, rounded corners=3pt, line width=0.3mm, fill=blue!10, draw=blue!80, font=\footnotesize},
    demographic_work/.style =    {align=center, text width=10cm, rounded corners=3pt, line width=0.3mm, fill=blue!10, draw=blue!0, font=\footnotesize},
    character/.style =         {align=center, text width=1.8cm, rounded corners=3pt, line width=0.3mm, fill=red!10, draw=red!80, font=\footnotesize},
    character_work/.style =    {align=center, text width=10cm, rounded corners=3pt, line width=0.3mm, fill=red!10, draw=red!0, font=\footnotesize},
    personalization/.style =           {align=center, text width=1.8cm, rounded corners=3pt, line width=0.3mm, fill=cyan!10, draw=cyan!80, font=\footnotesize},
    personalization_work/.style =      {align=center, text width=10cm, rounded corners=3pt, line width=0.3mm, fill=cyan!10, draw=cyan!0, font=\footnotesize},
    risk/.style =         {align=center, text width=1.8cm, rounded corners=3pt, line width=0.3mm, fill=orange!10, draw=orange!80, font=\footnotesize},
    risk_work/.style =    {align=center, text width=10cm, rounded corners=3pt, line width=0.3mm, fill=orange!10, draw=orange!0, font=\footnotesize},
}

\usepackage{CJK}

\title{LA4VLA: Learning to Act without Seeing 
\\via Language-Action Pretraining}

\author[1,2,*]{Tao Lin}
\author[1,2,*]{Yuxin Du}
\author[1,*]{Yiran Mao}
\author[1]{Zewei Ye}
\author[1]{Yilei Zhong}
\author[1]{Bing Cheng}
\author[1]{Yiming Wang}
\author[1]{\\Jiting Liu}
\author[1]{Yang Tian}
\author[1]{Junchi Yan}
\author[2]{Feiran Wu}
\author[2]{Zenan Meng}
\author[2]{Hu Wei}
\author[4,\dagger]{\\Yuqian Fu}
\author[3,\dagger]{Gen Li}
\author[1,\dagger]{Bo Zhao}

\affiliation[1]{School of AI, Shanghai Jiao Tong University}
\affiliation[2]{Alibaba Group\\}
\affiliation[3]{Nanyang Technological University}
\affiliation[4]{KAUST}

\contribution[*]{Equal contribution}
\contribution[\dagger]{Corresponding authors.}
\checkdata[Email]{taolin200108@gmail.com, bo.zhao@sjtu.edu.cn}
\checkdata[Project Page]{\url{https://github.com/MINT-SJTU/LA4VLA}}

\abstract{
Vision-Language-Action (VLA) models are commonly pretrained on robot demonstrations by jointly mapping visual observations and language instructions to actions. 
However, dense visual-action supervision can dominate the comparatively sparse language-action signal.
As a result, policies may rely on visual shortcuts rather than learn how language conditions action execution, making them sensitive to visual variations.
To address this limitation, we propose LA4VLA, a language-action pretraining framework that enables policies to acquire language-conditioned action priors without visual observations.
These priors capture reusable manipulation skills shared across tasks and scenes, reducing reliance on scene-specific visual cues.
Specifically, LA4VLA decomposes expert demonstration trajectories into atomic action segments and pairs each segment with a corresponding low-level action description. This yields LA-33K, a dataset of 33K Language-Action (LA) episodes derived entirely from existing demonstrations without additional robot data collection.
We further develop LA4VLA-1B, a lightweight 1B-parameter VLA model, and investigate three paradigms for incorporating language-action supervision into VLA learning: LA-only pretraining, sequential LA-to-VLA pretraining, and mixed LA-VLA pretraining.
Across simulation and real-world tasks, LA-pretrained policies consistently outperform matched VLA-pretrained counterparts, while combining LA and VLA supervision leads to further gains.
In particular, mixed LA-VLA pretraining improves the average success rate of LA4VLA-1B over the no-pretraining baseline by up to 17.8 and 45.0 percentage points in simulation and real-world tasks, respectively.
These results establish LA4VLA as an effective and complementary pretraining strategy for building stronger and more robust VLA policies.
}

\begin{document}

\maketitle

\section{Introduction}

\begin{figure*}[t]
  \centering
  \includegraphics[width=\textwidth]{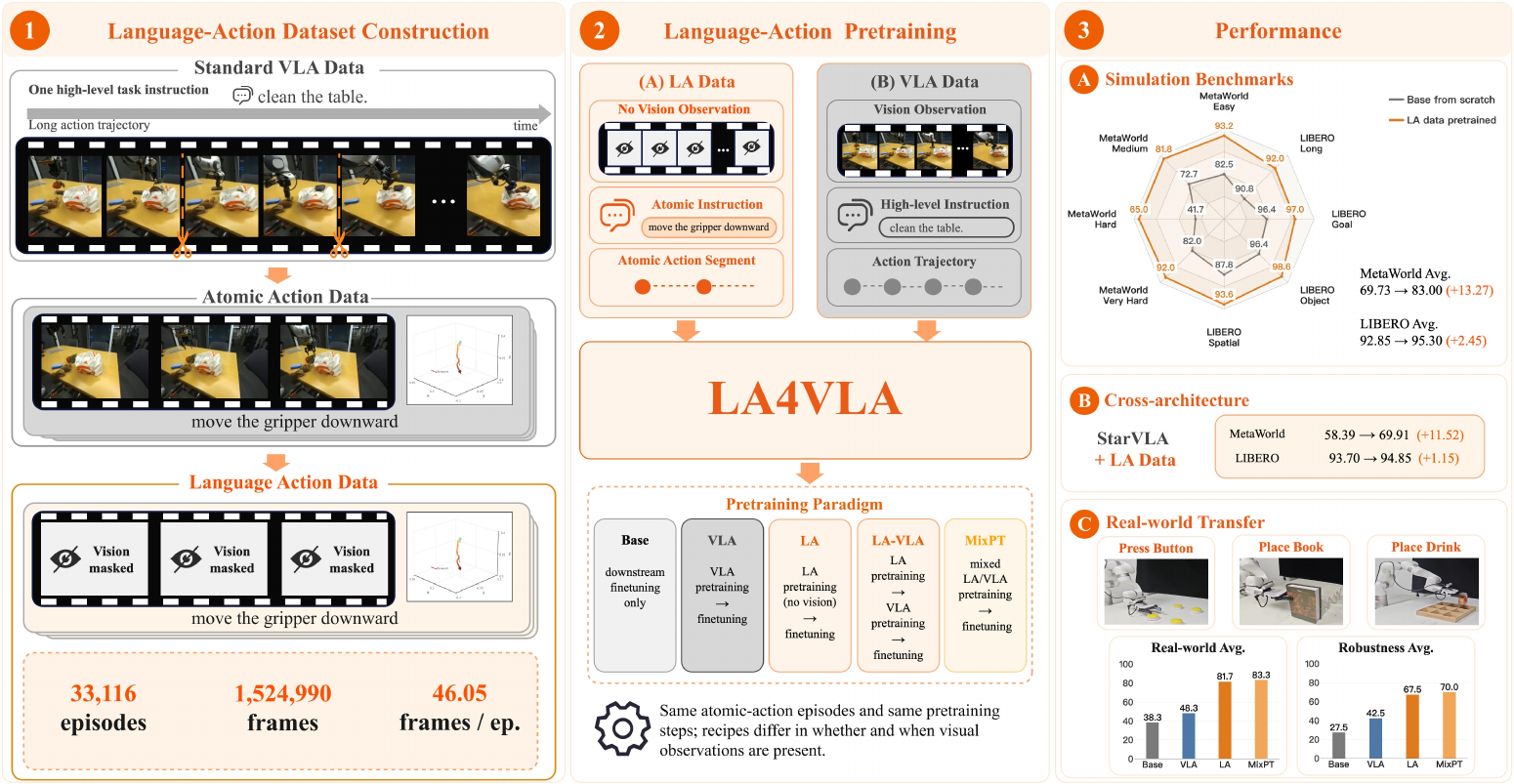}
\caption{
Overview of the LA4VLA framework.
The left panel illustrates the construction of Language-Action data by reorganizing VLA demonstrations into atomic instruction-action segments.
The middle panel summarizes the LA4VLA pretraining paradigm, which studies how LA supervision can be used alone or combined with standard VLA pretraining.
The right panel presents representative performance across simulation, architectures, real-world transfer, and robustness evaluations.
}
  \label{fig:main_overview}
\end{figure*}

Vision-Language-Action (VLA) models~\cite{kim2024openvla,black2410pi0,bjorck2025gr00t,wang2026qwen,lin2025evo,lin2026evodepth, wang2026afford, wang2026oflow,zuo2026gaze2act} have recently emerged as a promising paradigm for general-purpose robot manipulation. These models are typically trained on large-scale robot demonstrations~\cite{khazatsky2024droid,o2024open,walke2023bridgedata,fang2026molmoact2}, where each trajectory contains visual observations, action sequences, and a high-level language instruction specifying the task goal. By learning to predict actions from the current visual observations and the given instruction, this multimodal joint training paradigm enables robots to perform diverse manipulation tasks from language commands.

Despite its effectiveness, the joint VLA training paradigm introduces an inherent asymmetry in how vision and language contribute to action learning. This asymmetry appears at two levels. First, \textbf{at the data level}, a single robot trajectory can contain hundreds or even thousands of image-action pairs, but is typically paired with only one high-level task instruction. 
The instruction often specifies only the overall task goal, while visual observations and robot states vary continuously throughout the trajectory.
As a result, vision-action supervision is temporally dense and dynamically changing, whereas language-action supervision has much less semantic variation and lacks explicit alignment to local action stages. 
Second, \textbf{at the model-input level}, visual observations are usually encoded into hundreds of image tokens, often outnumbering the tokens of language instructions by one to two orders of magnitude.
Together, these data-level and input-level asymmetries can bias action learning toward visual signals and dilute the supervision for learning how language conditions actions. 
Consequently, a policy may appear language-conditioned, while its internal representations remain dominated by visual-action associations. 
This can lead to visual shortcuts that weaken the influence of language intent, leaving language-action grounding underdeveloped~\cite{xu2025seeing,fang2026vision}.
Such entanglement becomes especially problematic under visual variations, such as changes in viewpoint, background, lighting, object appearance, or scene layout, where robust manipulation requires actions to remain conditioned on the instruction rather than on spurious visual correlations.

This motivates us to rethink VLA pretraining through one key question: \textit{\textbf{Can language-action grounding be decoupled from visually dominated VLA learning?}}
We hypothesize that explicitly removing visual observations during pretraining can prevent the policy from exploiting visual shortcuts and encourage it to focus on the intrinsic relationship between language and action.
To this end, we explore vision-agnostic language-action pretraining, where the policy predicts actions from language instructions and proprioceptive states without visual observations. The objective is not intended to replace visual grounding, since object localization and scene-specific target selection still require vision. Instead, it encourages the policy to learn vision-independent action knowledge, such as action dynamics, temporal consistency, and language-conditioned motion patterns. 

Building on this idea, we propose \textbf{LA4VLA}, a language-action pretraining framework that constructs explicit Language-Action (LA) supervision and integrates it into VLA learning. 
Specifically, LA4VLA decomposes complete demonstration trajectories into short atomic action segments and associates each segment with a corresponding low-level action description. 
The resulting LA pairs enable the policy to learn how language conditions action execution without relying on visual observations. 
These atomic actions, such as moving, grasping, lifting, transporting, pressing, and rotating, capture reusable manipulation skills shared across tasks and scenes. 
Based on this trajectory decomposition, we build \textbf{LA-33K}, a dataset containing 33K LA pairs derived entirely from existing expert trajectories~\cite{khazatsky2024droid}, without requiring additional robot data collection. Rather than merely scaling the amount of training data, LA4VLA reorganizes existing demonstrations to expose explicit language-action supervision that is often implicit or diluted in standard VLA training. Pretraining on these vision-agnostic LA pairs enables the policy to acquire reusable language-conditioned action priors that can strengthen downstream VLA learning.

We systematically evaluate the effectiveness of LA pretraining by developing \textbf{LA4VLA-1B}, a lightweight 1B-parameter VLA model, and testing it across several simulation benchmarks and real-world robot manipulation tasks.
To examine how language-action supervision can be integrated into VLA policy learning, we study a series of pretraining paradigms, including pure LA pretraining, sequential LA-to-VLA pretraining, and mixed LA-VLA pretraining. Consistent results across LA4VLA-1B and StarVLA~\cite{community2026starvla} show that pretraining on LA data alone yields stronger downstream improvements than pretraining with VLA data alone, while combining LA and VLA data leads to further gains. These findings demonstrate that language-action supervision provides an effective and transferable training signal for downstream VLA policies and is complementary to existing VLA learning pipelines. 
In particular, LA4VLA-1B achieves average success rates of 87.53\% on MetaWorld~\cite{yu2020meta}, 96.28\% on LIBERO~\cite{liu2023libero}, and 83.3\% on real-world manipulation tasks, improving over the baselines by 17.80, 3.43, and 45.0 percentage points, respectively.
Beyond average task success, our analysis further shows that LA pretraining improves robustness under visual perturbations, supporting the hypothesis that language-conditioned action priors can reduce over-reliance on visual shortcuts. Figure~\ref{fig:main_overview} provides an overview of the LA4VLA framework, including LA data construction, the pretraining paradigm, and representative performance across simulation benchmarks, architectures, and real-world manipulation tasks. Overall, our main contributions are summarized as follows:

\begin{itemize}

\item We propose LA4VLA, a language-action pretraining framework that decouples language-conditioned action learning from visually grounded VLA training, enabling policies to learn reusable action priors that generalize across scenes and tasks without over-relying on visual cues.

\item We construct Language-Action (LA) data from existing expert demonstrations by decomposing full trajectories into atomic action segments and pairing each segment with a low-level action description. This results in LA-33K, a dataset of 33K vision-agnostic LA episodes created without additional robot data collection.

\item We show that LA pretraining consistently enhances VLA policy performance across simulation benchmarks, model architectures, and real-world manipulation tasks, improves robustness under visual perturbations, and yields further gains when combined with standard VLA data.

\end{itemize}

\section{Related Work}

\subsection{Vision-Language-Action Models and Robot Pretraining}

Vision-Language-Action (VLA) models have become a common paradigm for learning robot policies from language-conditioned demonstrations. Early robot transformer models showed that large-scale behavior cloning can support multi-task manipulation from visual observations and language instructions~\cite{brohan2022rt,zitkovich2023rt,o2024open}. Recent VLA models further scale this paradigm with stronger vision-language backbones, larger robot datasets, and more expressive action heads, leading to open-source and generalist robot policies such as OpenVLA, $\pi_0$, GR00T N1, SmolVLA, Evo-1, StarVLA, and Qwen-VLA~\cite{kim2024openvla,black2410pi0,bjorck2025gr00t,shukor2025smolvla,lin2026evo,community2026starvla,wang2026qwen}. These models have demonstrated strong performance across simulation benchmarks and real-world manipulation tasks, making VLA pretraining a central direction for general-purpose robot policy learning.

Despite differences in architectures, action representations, and training recipes, most existing VLA models share the same supervision format: the policy is pretrained on demonstrations that jointly pair visual observations, language instructions, and action trajectories. This joint formulation is effective for learning visually grounded robot policies, but it also couples visual grounding and action learning from the beginning. Recent work has improved this paradigm through better action tokenization, diffusion or flow-matching action heads, larger multi-robot datasets, and cross-embodiment conditioning~\cite{pertsch2025fast,black2410pi0,lin2026evodepth,wang2026qwen,a2a2026,du2026focusable,liu2025vla}. However, these improvements still primarily operate within vision-language-action joint training. As a result, the contribution of language-action supervision itself remains difficult to isolate, and it is unclear whether the policy learns a reusable instruction-conditioned action prior or mainly uses language as a weak condition alongside dense visual observations. 
In contrast, our work explicitly separates language-action learning from visual grounding during pretraining, allowing us to study how vision-agnostic action supervision can benefit VLA policies.

\subsection{Language and Action Supervision for Robot Learning}

Language has long been used to specify robot tasks, goals, and intermediate plans. Language-conditioned imitation learning and instruction-following manipulation train policies to execute natural-language commands from visual observations~\cite{jang2022bc,lynch2023interactive, shridhar2022cliport}. Other systems use language models to decompose tasks, generate plans, or provide reasoning traces for robot control~\cite{ahn2022can,liang2023code,zhao2025cot,shi2025hi,li2026mask2iv,mon2025embodied}. These approaches show that language is a useful interface for robot learning, but in many cases language serves as a task description, high-level plan, or reasoning scaffold rather than a direct source of action supervision.

More recent work has introduced language-to-action signals more explicitly. LAP represents low-level robot motions as natural-language action strings, such as descriptions of translational or rotational end-effector displacement, and trains a VLM backbone to predict them~\cite{zha2026lap}, while Qwen-VLA includes a text-to-action stage that trains an action decoder from language and embodiment prompts~\cite{wang2026qwen}.
These studies demonstrate that language can provide more direct supervision for action learning, but do not investigate vision-agnostic language-action supervision as a pretraining signal for VLA policy learning.
In contrast, our work pairs atomic instructions with their corresponding action trajectories, removes visual observations during LA pretraining, and systematically studies how this supervision can be used alone or combined with standard VLA pretraining.

\section{Probing Instruction Following in VLAs}
\label{sec:visual_cue_dependence}

To examine whether a VLA policy predicts actions by following the instruction or by exploiting the visual observations paired with that instruction, we design a controlled diagnostic study. Taking the standard paired VLA input as the reference, we keep the language instruction fixed while removing, mismatching, or conflicting the visual input, and then systematically examine how the predicted action trajectory changes. Our assumption is that, if a policy has learned a robust language-action relation, fixing the instruction should keep the predicted motion aligned with the commanded direction. Conversely, if the predicted direction changes with the substituted visual input while the instruction remains fixed, it suggests that the policy may rely on visual cues rather than on language-action grounding. Instead of conducting this diagnostic on full trajectories, we focus on atomic direction-following actions, such as ``lower the object downward toward the target'' and ``transport the object horizontally to the left'', where each instruction specifies a short local motion and the expected action direction is directly measurable.

\subsection{Diagnostic Evaluation}

\noindent \textbf{Policy and task setup.}
For this diagnostic evaluation, we use a VLA policy built on InternVL3-1B~\cite{zhu2025internvl3} with a flow matching action head for continuous action prediction.
It is trained under the standard VLA format, i.e., with paired visual observations, language instructions, robot states, and action trajectories. The only difference from standard VLA training is that the policy is trained and evaluated on atomic direction-following trajectories, where the end effector moves along a specified Cartesian direction.
We construct direction-following cases from pairs of atomic-action instructions that differ only in direction, such as motions along the positive and negative directions of the same Cartesian axis. For each instruction, the policy predicts an action trajectory.

\noindent \textbf{Evaluated input conditions.}
We evaluate four visual-input configurations while keeping the target instruction \(L\) and robot state fixed. Specifically, we consider: 1) \textit{Standard paired input} \((L,V^+)\), where \(V^+\) denotes the visual observation originally paired with the instruction in standard VLA training; 2) \textit{Visual-removed input} \((L,\varnothing)\), where the visual observation is masked out; 3) \textit{Visual-unaligned input} \((L,\tilde{V})\), where the visual observation is replaced with an same-scene observation that is not paired with the current instruction; and 4) \textit{Visual-conflict input} \((L,V^-)\), where the visual observation is sampled from a trajectory associated with the opposite motion direction. 
Since the instruction and robot state remain fixed across all conditions, changes in the predicted action direction can be attributed to the visual input, allowing us to assess whether the policy follows the language instruction or relies on the paired visual observation.

\noindent \textbf{Metrics.}
We also use four metrics to quantify whether the predicted motion follows the instruction. The first two metrics measure directional alignment. \textit{Directional Alignment Rate} (DAR) is the fraction of trials whose endpoint displacement falls in the instruction-aligned half-space; $0.5$ corresponds to random directional choice, and values below $0.5$ indicate that predictions more often fall in the opposite half-space. \textit{Direction Consistency Score} (DCS) is the cosine similarity between the predicted endpoint displacement and the ground-truth endpoint displacement. Since the ground-truth displacement is collected under the corresponding directional instruction, DCS evaluates whether the predicted motion follows the instruction-conditioned direction, ranging from $-1$ to $1$, where larger values indicate stronger alignment with the demonstrated motion direction.
The other two metrics measure whether instructions specifying different target directions produce separable motion patterns.  \textit{Separability Ratio} (SR) compares the average distance across different instruction directions with the average distance within the same instruction direction. \textit{Silhouette Score} (SS) provides a normalized sample-wise clustering measure, evaluating whether each endpoint displacement is closer to its own instruction-direction cluster than to other direction clusters. Higher SR and SS indicate clearer instruction-conditioned separation. Detailed definitions and formulas for all four metrics are provided in Appendix~\ref{app:metrics_for_instruction_analysis}.

\begin{figure*}[t]
  \centering
  \includegraphics[width=\textwidth]{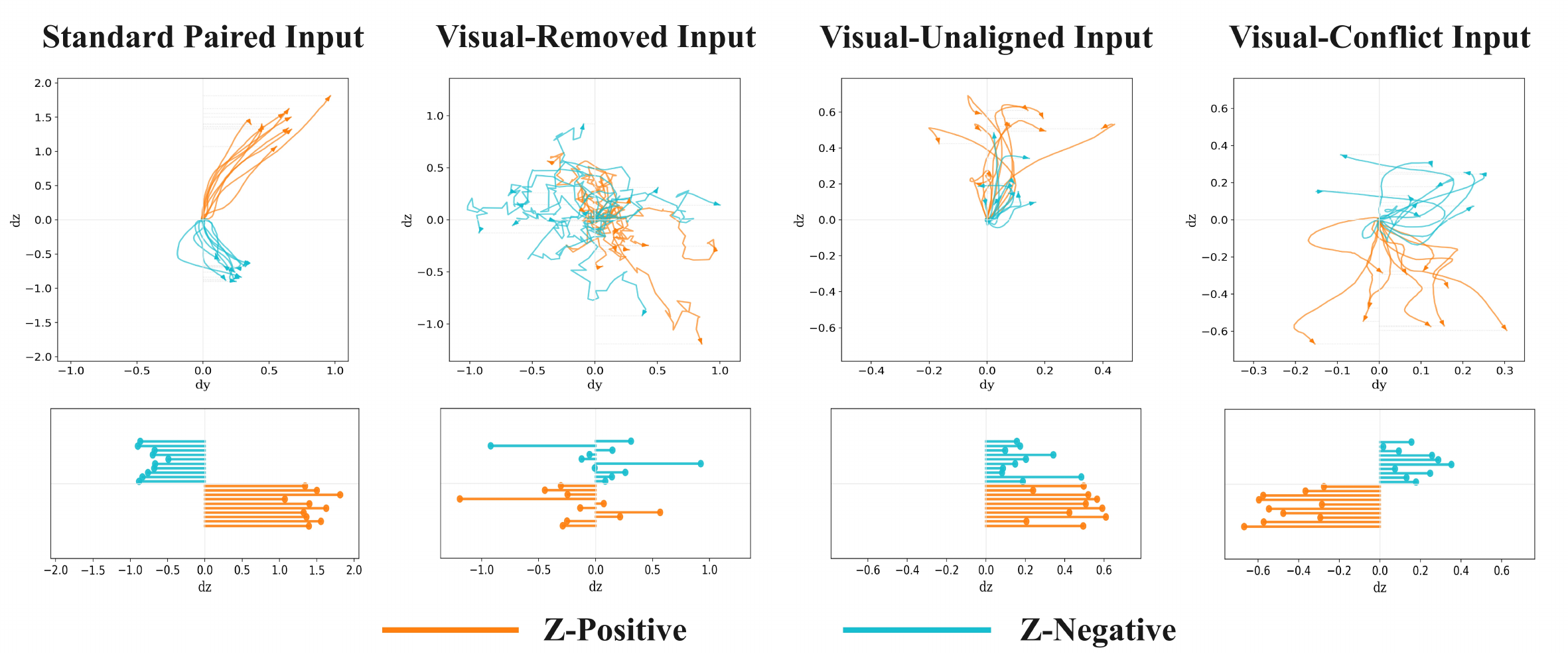}
\caption{
Qualitative visualization of instruction-conditioned direction following under different visual input configurations.
The top row shows predicted trajectories projected onto the plane containing the tested motion axis, and the bottom row presents endpoint displacement projected onto that axis.
}
  \label{fig:visual_cue_dependence}
\end{figure*}

\begin{table}[t]
\centering
\caption{
Instruction-following diagnostic under different visual input configurations.
Metrics are computed over 100 direction-following cases for each evaluation condition.
}
\label{tab:visual_cue_dependence}
\begin{tabular}{llcccc}
\toprule
Condition & Inputs & DAR $\uparrow$ & DCS $\uparrow$ & SR $\uparrow$ & SS $\uparrow$ \\
\midrule
Standard paired input & $(L, V^+)$ & 0.98 & 0.95 & 2.35 & 0.55 \\
Visual-removed input & $(L, \varnothing)$ & 0.63 & 0.16 & 1.03 & 0.04 \\
Visual-unaligned input & $(L, \tilde{V})$ & 0.66 & 0.37 & 1.13 & 0.05 \\
Visual-conflict input & $(L, V^-)$ & 0.35 & 0.03 & 1.03 & -0.04 \\
\bottomrule
\end{tabular}
\end{table}

\subsection{Diagnostic Results}

\noindent \textbf{Qualitative visualization.}
Figure \ref{fig:visual_cue_dependence} visualizes the predicted trajectories for a representative pair of instructions specifying opposite motion directions: ``move upward to approach the target'' and ``move downward to approach the target''.
For each prediction, we compute the start-to-end displacement vector and project both the complete trajectory and displacement vector onto the plane containing the tested motion axis.
Under the \textit{Standard paired input}, the model produces clearly separated trajectories for the two instructions, with endpoint projections falling on the instruction-aligned sides of the tested axis. However, the following visual interventions reveal that this alignment is fragile and not robustly grounded in the language instruction itself.
With the \textit{Visual-removed input}, the trajectories become scattered and the two instruction directions are poorly separated. With the \textit{Visual-unaligned input}, opposite instructions no longer produce clearly separable endpoint distributions, indicating that the policy cannot reliably distinguish actions based on language once the paired visual observation is replaced. Under the \textit{Visual-conflict input}, the substituted visual observation is associated with the opposite motion direction. In this case, the endpoint projections shift away from the instruction-aligned direction and become biased toward the direction implied by the substituted visual observation. 
These results show that the policy is highly sensitive to visual input and may deviate from the language instruction when the paired visual cue changes.

\noindent \textbf{Quantitative analysis.}
Table \ref{tab:visual_cue_dependence} reports the average results over 100 diverse direction-following cases for each evaluation condition. Under the \textit{Standard paired input}, the model appears to follow the instruction, reaching DAR of $0.98$ and DCS of $0.95$, with clear endpoint separation between opposite instructions (SR $2.35$, SS $0.55$). However, this apparent alignment quickly breaks under visual intervention. With the \textit{Visual-removed input}, DCS drops to $0.16$ and SS becomes nearly zero, indicating that opposite instructions no longer induce clearly separable actions without the paired image. With the \textit{Visual-unaligned input}, the separation remains poor, with DCS of $0.37$ and SS of $0.05$, showing that replacing the paired image with an unaligned same-scene observation is enough to disrupt instruction-conditioned action prediction.
The \textit{Visual-conflict input} gives the strongest evidence: DAR drops to $0.35$, well below the $0.5$ expected under random directional choice, and DCS falls to $0.03$, meaning that predictions fall in the opposite half-space more often than in the instruction-aligned half-space. This indicates that the policy is not merely uncertain, but can be biased away from the instruction by a conflicting visual observation. 

These quantitative results are consistent with the qualitative visualizations in Figure~\ref{fig:visual_cue_dependence}: standard paired inputs can create apparent instruction following, but once the paired visual cue is removed, unaligned, or placed in conflict, the policy fails to reliably distinguish actions according to language. Since the instruction is fixed across these conditions, this behavior directly supports our diagnostic assumption that changes in predicted motion reflect the policy's dependence on the substituted visual input. 
Together, the qualitative and quantitative results suggest that standard VLA training does not robustly ground action execution in language, motivating more direct language-action supervision without paired visual observations.

\begin{figure}[ht]
\centering
\includegraphics[width=0.98\linewidth]{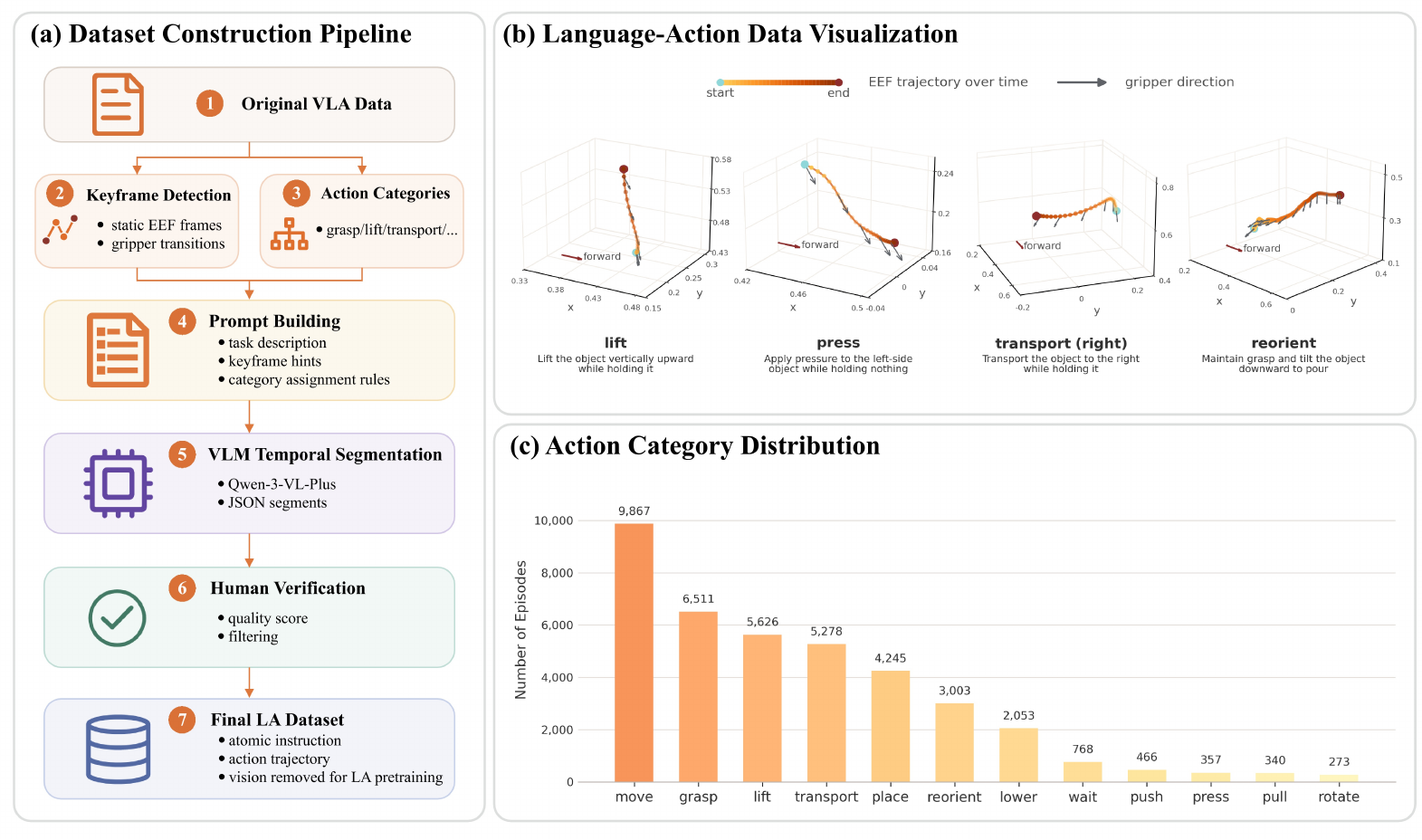}
\caption{
Overview of the Language-Action dataset.
The left panel summarizes the construction pipeline, which converts standard vision-language-action demonstrations into vision-agnostic atomic language-action episodes. The top-right panel visualizes representative Language-Action data examples using end-effector trajectories and gripper-direction arrows.
The bottom-right panel reports the action-category distribution of the LA-33K dataset.
}
\label{fig:la_dataset_overview}
\end{figure}

\section{LA4VLA: Language-Action Dataset Construction}
\label{sec:la_dataset}

\subsection{Overview}
The diagnostic analysis above shows that the instruction-following behavior of VLA policies can become unstable under changes in visual input. This suggests that language-action supervision should be exposed more directly, rather than always being tied to paired visual observations. 
To this end, we propose the LA4VLA framework for vision-agnostic language-action pretraining, where models learn from language instructions, proprioceptive states, and action trajectories without visual inputs.

To convert standard VLA demonstrations into LA episodes, LA4VLA introduces a construction pipeline that segments long demonstration trajectories into short atomic action intervals and annotates each interval with a low-level instruction. Visual observations are used only during construction and verification, while the resulting LA episodes remove visual inputs and retain only the instruction, proprioceptive states, and corresponding action trajectory for pretraining.

We apply this pipeline to a large-scale robot manipulation dataset~\cite{khazatsky2024droid}, resulting in LA-33K, a dataset of 33K Language-Action episodes constructed without additional robot data collection. By reorganizing standard VLA demonstrations into explicit instruction-action pairs, LA-33K exposes language-action supervision that is often implicit or diluted in full demonstration trajectories. Figure~\ref{fig:la_dataset_overview} summarizes the construction pipeline, visualized examples of LA data, and the action-category distribution of LA-33K.

\subsection{Dataset Construction Pipeline}

\noindent \textbf{Pipeline overview.}
As shown in Figure~\ref{fig:la_dataset_overview}(a), it illustrates how the original VLA data, consisting of visual observations, robot states, language instructions, and action trajectories, is reorganized into a Language-Action dataset of vision-agnostic atomic language-action episodes.
For each demonstration, we first extract keyframe hints from robot states and define an atomic-action vocabulary that constrains the candidate action categories. These two sources of information are then combined with the original task description to build a structured prompt, which asks a VLM to propose temporal segments with primitive labels, vision-agnostic instructions, and temporal boundaries. The VLM-generated candidates are subsequently inspected through human verification, where low-quality or misaligned segments are filtered out. The retained segments form the final LA dataset, where each episode contains an atomic instruction, robot states, and the corresponding action trajectory, with visual observations removed for LA pretraining.

\noindent \textbf{Keyframe detection.}
We use temporal cues from the robot state trajectory to guide segmentation. 
Static keyframes mark short intervals with low arm-state variation and often correspond to alignment, grasping, placing, or phase transitions. Gripper-transition keyframes mark frames where the gripper changes between open and closed states, which commonly indicate grasp or release phases. These detected events are inserted into the segmentation prompt as timestamped hints. They are used as soft cues rather than fixed boundaries, allowing the segmentation model to combine robot-state signals with visual evidence when proposing atomic segments.

\noindent \textbf{Action categories.}
In parallel with keyframe detection, we define an atomic-action vocabulary that constrains the segmentation output. As summarized in Table~\ref{tab:atomic_taxonomy}, the vocabulary is organized by interaction context and motion type, covering common manipulation primitives such as \textit{grasp}, \textit{place}, \textit{lift}, \textit{lower}, \textit{transport}, \textit{push}, \textit{pull}, \textit{press}, \textit{move}, \textit{rotate}, and \textit{reorient}. Each category is defined by the dominant physical effect of the segment, using end-effector motion, gripper state, object status, and contact context as cues. These categories provide the structured labels used in the prompt and help the VLM assign each candidate segment to a consistent action type before generating the corresponding vision-agnostic instruction.

\begin{table}[ht]
\centering
\caption{Atomic action categories organized by interaction context and motion type.}
\label{tab:atomic_taxonomy}
\begin{tabular}{p{0.18\linewidth}p{0.26\linewidth}p{0.25\linewidth}p{0.20\linewidth}}
\toprule
 & Object manipulation & Contact interaction & Free-space motion \\
\midrule
Gripper & grasp / place & -- & -- \\
Translation & lift / lower / transport & push / pull / press & move \\
Rotation & wrist\_rotate / reorient & -- & rotate / reorient \\
\bottomrule
\end{tabular}
\end{table}

\noindent \textbf{Prompt construction.}
For each demonstration, we build a structured prompt that includes the original task instruction, sampled multi-view video frames, detected keyframe events, the atomic-action vocabulary, primitive definitions, category assignment rules, and a required JSON output schema. The prompt asks the VLM to produce both a structured primitive label and a natural-language instruction for each segment. The instruction is written in a compact, vision-agnostic form, describing action type, direction, and optional object or object-state information while avoiding appearance and scene-specific details.

\noindent \textbf{VLM-based temporal segmentation.}
We use Qwen-3-VL-Plus~\cite{bai2025qwen3} as a proposal generator for atomic temporal segmentation. Video frames are sampled at $2$ FPS and resized to $224 \times 224$; when multiple camera views are available, each view is provided with a textual view label. Given the prompt and video inputs, the VLM outputs a JSON list of candidate segments, where each segment contains a primitive label, a vision-agnostic instruction, and start/end timestamps. 
In this stage, visual observations are used only to identify and describe candidate action segments during data construction, but are excluded from the resulting LA dataset.

\noindent \textbf{Human verification.}
The VLM-generated candidates are then filtered by human annotators. For each candidate segment, annotators inspect the corresponding video clip together with the primitive label and instruction, and evaluate whether the temporal boundary, action label, and language description are mutually consistent. Each segment receives a quality score from $0$ to $3$, and we retain segments with quality score at least $2$. This step removes noisy proposals such as over-segmented clips, merged primitives, inaccurate boundaries, and instructions that do not match the observed action. The retained segments form the final LA dataset used for language-action pretraining. Additional implementation details of the construction pipeline are provided in Appendix~\ref{app:dataset_construction}.

\subsection{Dataset Statistics}
Applying the construction pipeline to DROID~\cite{khazatsky2024droid} yields LA-33K. We next summarize its scale, granularity, verification quality, and action coverage.

\noindent \textbf{Scale and granularity.}
Table~\ref{tab:la_dataset_statistics} summarizes how the data changes across processing phases during LA dataset construction. Starting from $9{,}560$ original VLA episodes, VLM-based segmentation produces $56{,}899$ candidate atomic language-action episodes. After human verification, $33{,}116$ episodes are retained as the final LA dataset. Compared with the original vision-language-action episodes, the final LA episodes are much shorter on average, reducing the average number of frames per episode from $287.83$ to $46.05$. This makes language supervision temporally denser: instead of assigning one high-level instruction to a long visual trajectory, LA-33K pairs short action segments with localized atomic instructions. In this way, LA-33K converts sparse task-level language supervision into dense, localized language-action supervision that is more directly aligned with action execution.

\begin{table}[t]
\centering
\small
\caption{Dataset scale and episode granularity during LA dataset construction.}
\label{tab:la_dataset_statistics}
\begin{tabular}{lrrr}
\toprule
Processing Phase & Episodes & Frames & Frames / Ep. \\
\midrule
Original VLA episodes & 9,560 & 2,751,615 & 287.83 \\
VLM-generated LA candidates & 56,899 & 2,551,102 & 44.84 \\
Final human-verified LA episodes & 33,116 & 1,524,990 & 46.05 \\
\bottomrule
\end{tabular}
\end{table}

\noindent \textbf{Verification quality.}
VLM-based segmentation produces $5.95$ candidate atomic language-action episodes per original VLA episode on average. 
We retain segments that pass the predefined quality threshold, resulting in a candidate retention rate of $58.2\%$. Compared with the original $9{,}560$ VLA episodes, the final LA dataset contains $3.46$ times as many episodes. On a doubly annotated subset, inter-annotator agreement reaches Krippendorff's $\alpha = 0.7794$, indicating substantial consistency in the verification process. The full annotation and filtering protocol is described in Appendix~\ref{app:dataset_construction}.

\noindent \textbf{Action coverage.}
Figure~\ref{fig:la_dataset_overview}(c) reports the action-category distribution of LA-33K. The final dataset covers frequent object-manipulation and free-space categories such as \textit{move}, \textit{grasp}, \textit{lift}, \textit{transport}, and \textit{place}, while also retaining contact-interaction and rotational categories such as \textit{push}, \textit{press}, \textit{pull}, \textit{rotate}, and \textit{reorient}. Overall, the distribution indicates that LA-33K provides broad coverage of atomic manipulation behaviors for language-action pretraining.

\section{LA4VLA: Language-Action Pretraining}

\subsection{Experimental Designs}

This section systematically evaluates language-action pretraining as a VLA learning paradigm, examining its downstream performance, integration with VLA pretraining, transferability, robustness, and effects on instruction-conditioned action prediction.

\noindent \textbf{Pretraining paradigms.}
We study a pretraining paradigm built around vision-agnostic language-action supervision for VLA policy learning. We use two paired pretraining data formats derived from the same atomic episodes. The first is LA-33K, where visual observations are removed and each episode contains a vision-agnostic instruction, robot states, and the corresponding action trajectory. The second is LA-33K-V, a VLA-format counterpart of LA-33K that restores the corresponding original visual observations for each atomic episode. \textbf{Base} directly finetunes the policy on downstream task data without pretraining. \textbf{LA} pretrains on LA-33K without visual input before the same downstream finetuning stage. \textbf{VLA} pretrains on LA-33K-V with visual observations enabled, and then finetunes on the downstream task. \textbf{LA-VLA} first performs LA pretraining on LA-33K, then performs VLA pretraining on LA-33K-V, and finally finetunes on downstream task data. \textbf{MixPT} jointly uses LA-33K and LA-33K-V within a single mixed pretraining stage, followed by the same downstream finetuning.

\noindent \textbf{Questions and findings.}
The experiments address six questions:
\begin{itemize}
    \item \textbf{Does language-action pretraining improve downstream performance?} With identical downstream finetuning, LA-pretrained policies consistently achieve higher success rates than policies trained without pretraining or with VLA pretraining (Section~\ref{sec:exp_downstream}).
    \item \textbf{How should LA supervision be combined with VLA pretraining?} We compare pretraining only with language-action supervision (LA), pretraining first with language-action supervision and then with visual observations enabled (LA-VLA), and mixing language-action and vision-language-action data within one pretraining stage (MixPT). The results show that incorporating visual grounding after or together with language-action pretraining further strengthens downstream VLA policies (Section~\ref{sec:exp_recipes}).
    
    \item \textbf{Does the benefit generalize across VLA architectures?} We evaluate LA pretraining on another VLA architecture and observe consistent downstream gains, showing that language-action supervision can transfer across model architectures (Section~\ref{sec:exp_cross_arch}).
    
    \item \textbf{Does the benefit extend to to real robots?} On real-world language-conditioned manipulation tasks, LA pretraining leads to stronger real-world manipulation capability (Section~\ref{sec:exp_realworld}).
    
    \item \textbf{Is the benefit robust to visual noise?} Under visual perturbations, LA pretraining improves the policy's robustness and preserves stronger manipulation capability when observations are degraded (Section~\ref{sec:exp_generalization}).
    
    \item \textbf{Do the predicted actions follow the language instruction?} Under controlled conflict settings, the LA-pretrained policy still predicts action directions that follow the instruction, and its action-decoding representations show clearer instruction-conditioned separation (Section~\ref{sec:exp_behavior}).

\end{itemize}

\noindent \textbf{Experimental setup.}
We propose \textbf{LA4VLA-1B}, a 1B-parameter VLA model built on InternVL3-1B with a flow matching action head. We pretrain the model under the different paradigms described above.
For LA-format batches, visual observations are masked out and the model predicts action trajectories from language instructions and robot states. For VLA-format batches and downstream finetuning, visual observations are enabled. For fair comparisons within each experiment group, we keep the model architecture, training budget, optimization hyperparameters, downstream data, and finetuning protocol fixed, changing only the pretraining paradigm under study.

\subsection{Effect of Language-Action Pretraining}
\label{sec:exp_downstream}

We first evaluate whether language-action pretraining improves downstream manipulation performance. Table~\ref{tab:downstream_performance} reports results on MetaWorld~\cite{yu2020meta} and LIBERO~\cite{liu2023libero}. Results of prior methods are taken from their original papers or official reproductions reported in prior work.

On MetaWorld, LA pretraining improves over both from-scratch finetuning and VLA pretraining, reaching $83.00\%$ average success compared with $79.78\%$ for VLA pretraining and $69.73\%$ for the Base policy. On LIBERO, the same ordering holds: LA pretraining reaches $95.30\%$ average success, above the VLA-pretrained policy at $94.40\%$ and the Base policy at $92.85\%$, in a benchmark where overall success rates are already high. These results show that vision-agnostic language-action supervision is more effective than VLA pretraining under the matched atomic-action pretraining setting. The gains are larger on MetaWorld, where the benchmark leaves more room for improvement, and remain visible on LIBERO despite stronger baseline performance. We next ablate how language-action supervision should be integrated with VLA pretraining.

\begin{table*}[t]
\centering
\small
\begingroup
\renewcommand{\arraystretch}{0.88}
\setlength{\aboverulesep}{0.35ex}
\setlength{\belowrulesep}{0.35ex}
\caption{Downstream performance under different pretraining schemes. Success rates are reported in percent. The best and second-best values in each numeric column are marked in \textbf{bold} and \underline{underline}, respectively. Reference results are taken from original papers or official reproductions reported in existing work.}
\label{tab:downstream_performance}
\begin{tabular*}{\textwidth}{@{\extracolsep{\fill}}llcccccc@{}}
\toprule
{\bfseries Model} & {\bfseries Pretrain} & \multicolumn{5}{c}{\bfseries Success Rate (\%)} & {\bfseries Gain} \\
\midrule
{\bfseries MetaWorld} & & {\bfseries Easy} & {\bfseries Medium} & {\bfseries Hard} & {\bfseries Very Hard} & {\bfseries Avg.} & \\
\midrule
GR-1~\cite{wu2024unleashing} & VLA & 76.6 & 35.3 & 46.0 & 44.0 & 50.5 & -- \\
SmolVLA~\cite{shukor2025smolvla} & No & 87.1 & 51.8 & 70.0 & 64.0 & 68.2 & -- \\
RoboTron Mani~\cite{yan2024robotron} & VLA & 85.5 & 67.7 & 76.7 & 81.0 & 77.7 & -- \\
Evo-1~\cite{lin2026evo} & No & 89.2 & 76.8 & 77.2 & 79.2 & 80.6 & -- \\
$\pi_{0.5}$~\cite{physicalintelligence2025pi05} & Yes & 86.1 & 78.2 & 80.0 & 82.0 & 81.6 & -- \\
Evo-depth~\cite{lin2026evodepth} & No & 83.1 & 84.7 & \textbf{87.3} & 82.4 & 84.4 & -- \\
ALAM~\cite{tang2026alam} & VLA & 89.3 & 83.6 & \underline{85.0} & 82.0 & 85.0 & -- \\
\midrule
StarVLA~\cite{community2026starvla} & No & 85.7 & 60.0 & 40.6 & 47.3 & 58.39 & +0.00 \\
StarVLA~\cite{community2026starvla} & LA & 85.1 & 62.2 & 55.0 & 77.3 & 69.91 & +11.52 \\
\midrule
LA4VLA-1B & No & 82.5 & 72.7 & 41.7 & 82.0 & 69.73 & +0.00 \\
LA4VLA-1B & VLA & 88.6 & 68.2 & 68.3 & 94.0 & 79.78 & +10.05 \\
LA4VLA-1B & LA & \textbf{93.2} & 81.8 & 65.0 & 92.0 & 83.00 & +13.27 \\
LA4VLA-1B & LA-VLA & \underline{91.4} & \underline{90.9} & 66.7 & \underline{98.0} & \underline{86.75} & \underline{+17.02} \\
LA4VLA-1B & MixPT & 88.9 & \textbf{94.5} & 66.7 & \textbf{100.0} & \textbf{87.53} & \textbf{+17.80} \\
\midrule
{\bfseries LIBERO} & & {\bfseries Spatial} & {\bfseries Object} & {\bfseries Goal} & {\bfseries Long} & {\bfseries Avg.} & \\
\midrule
OpenVLA~\cite{kim2024openvla} & VLA & 84.7 & 88.4 & 79.2 & 53.7 & 76.5 & -- \\
$\pi_0$-FAST~\cite{pertsch2025fast} & VLA & 96.4 & 96.8 & 88.6 & 60.2 & 85.5 & -- \\
GR00T N1~\cite{bjorck2025gr00t} & VLA & 94.4 & 97.6 & 93.0 & 90.6 & 93.9 & -- \\
$\pi_0$~\cite{black2410pi0} & VLA & \underline{96.8} & 98.8 & 95.8 & 85.2 & 94.2 & -- \\
Evo-1~\cite{lin2026evo} & No & 92.7 & 97.7 & 96.3 & \underline{92.3} & 94.8 & -- \\
Evo-depth~\cite{lin2026evodepth} & No & 95.6 & \underline{99.2} & 95.6 & 91.3 & 95.4 & -- \\
\midrule
StarVLA~\cite{community2026starvla} & No & 96.4 & 98.4 & 95.8 & 84.2 & 93.70 & +0.00 \\
StarVLA~\cite{community2026starvla} & LA & \textbf{97.4} & 99.0 & 96.0 & 87.0 & 94.85 & +1.15 \\
\midrule
LA4VLA-1B & No & 87.8 & 96.4 & 96.4 & 90.8 & 92.85 & +0.00 \\
LA4VLA-1B & VLA & 93.2 & \textbf{99.4} & 96.8 & 88.2 & 94.40 & +1.55 \\
LA4VLA-1B & LA & 93.6 & 98.6 & 97.0 & 92.0 & 95.30 & +2.45 \\
LA4VLA-1B & MixPT & 93.4 & 98.2 & \underline{98.0} & \textbf{93.4} & \underline{95.75} & \underline{+2.90} \\
LA4VLA-1B & LA-VLA & 95.8 & 97.4 & \textbf{98.5} & \textbf{93.4} & \textbf{96.28} & \textbf{+3.43} \\
\bottomrule
\end{tabular*}
\endgroup
\end{table*}

\subsection{Integrating Language-Action Supervision with VLA Pretraining}
\label{sec:exp_recipes}

We further study how language-action supervision should be combined with VLA pretraining. Table~\ref{tab:downstream_performance} compares two pretraining paradigms: LA-VLA, which first performs LA pretraining and then VLA pretraining, and MixPT, which mixes LA and VLA data within a single pretraining stage. These two paradigms differ in how vision-agnostic language-action supervision and vision-language-action supervision are scheduled during pretraining.

On MetaWorld, both integration paradigms improve over LA pretraining alone. LA-VLA increases the average success rate from 83.00\% to 86.75\%, while MixPT further improves it to 87.53\%. On LIBERO, where baseline performance is already high, the same pattern remains: LA-VLA reaches 96.28\% and MixPT reaches 95.75\%, both above LA pretraining alone at 95.30\%. These results indicate that language-action supervision is not only effective as a standalone pretraining signal, but can also be complemented by VLA pretraining. Sequential LA-VLA pretraining gives the best result on LIBERO, while MixPT gives the best result on MetaWorld, suggesting that both staged and mixed integration are effective ways to combine language-action learning with visual grounding.

\subsection{Generalization to Different VLA Architectures}
\label{sec:exp_cross_arch}

We further investigate whether the benefits of LA pretraining transfer across VLA architectures. To this end, we apply the same LA pretraining protocol to StarVLA, another VLA architecture. We then finetune the LA-pretrained StarVLA and the no-pretraining StarVLA under the same downstream protocol and compare their final downstream performance, as shown in Table~\ref{tab:downstream_performance}.

The StarVLA results show that LA pretraining brings gains on both benchmarks. On MetaWorld, the average success rate increases from 58.39\% to 69.91\%, a gain of 11.52 points. The improvement is especially large on the hard and very hard buckets, where success increases by 14.4 and 30.0 points, respectively. On LIBERO, the average success rate increases from 93.70\% to 94.85\%, a gain of 1.15 points, with the largest improvement on the long-horizon suite. These results show that language-action supervision can transfer across VLA architectures and consistently benefit downstream manipulation performance.

\subsection{Real-World Manipulation Performance}
\label{sec:exp_realworld}

\begin{figure}[t]
\centering
\includegraphics[width=0.98\linewidth]{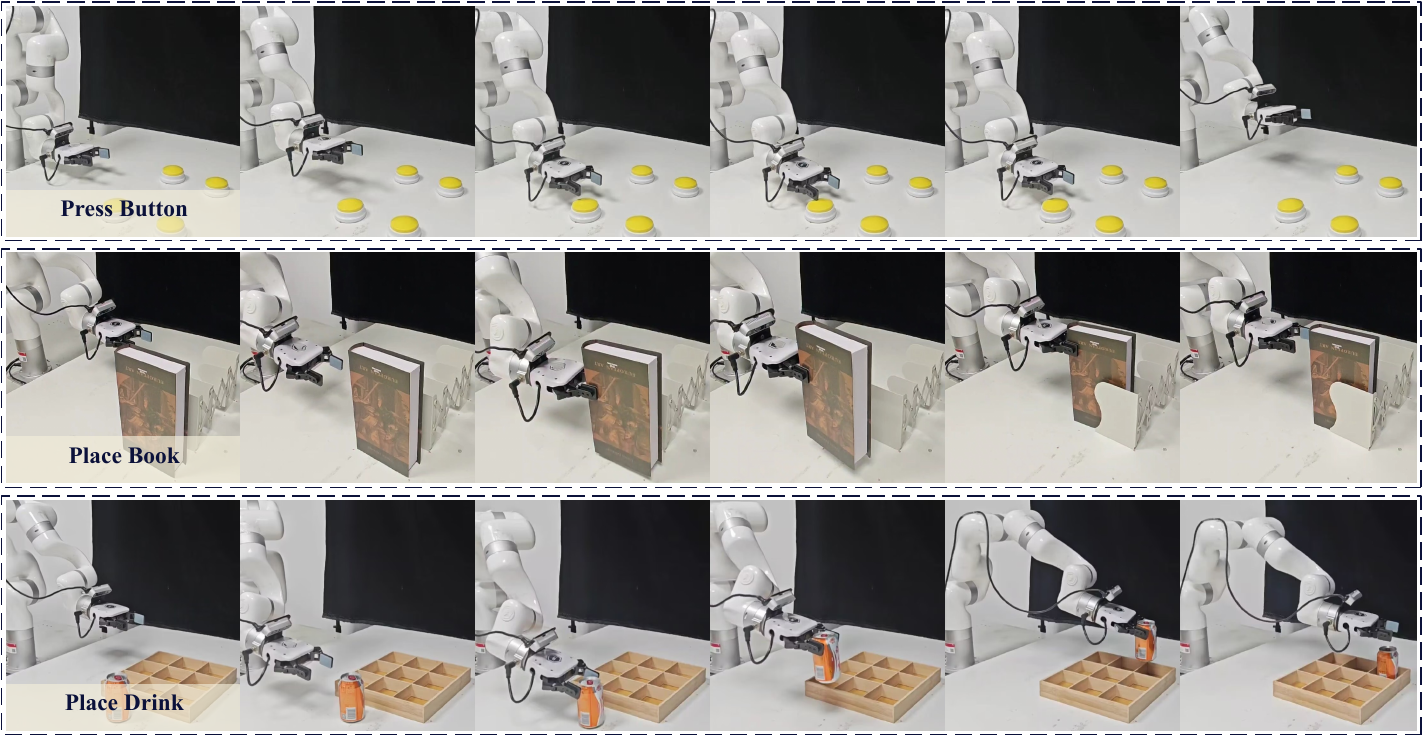}
\caption{Real-world task execution on the xArm6 platform. Each row shows a representative rollout of one language-conditioned task, with frames ordered from left to right over the course of execution: \textit{Press Button} (top), \textit{Place Book} (middle), and \textit{Place Drink} (bottom).}
\label{fig:real_world_process}
\end{figure}

We further evaluate whether the benefits of LA pretraining extend to real-world manipulation. The experiments use an xArm6 arm with a parallel-jaw gripper and two RGB cameras, one wrist-mounted and one third-person. We design three language-conditioned tasks: pressing one of four buttons placed at different positions on the table, inserting a book into one of three shelf positions, and placing a drink into one of nine grid cells. In each task, the target button, shelf position, or grid cell is specified by the language instruction. Figure~\ref{fig:real_world_process} shows representative rollouts of the three tasks.

Within each task, the initial scene is shared across target goals. Therefore, the visual observation alone does not specify which target should be selected; the language instruction determines the intended button, shelf position, or grid cell. For each pretraining scheme, we finetune one multi-task policy using demonstrations from all three tasks, with $100$ teleoperated demonstrations per task and $300$ demonstrations in total. The same finetuned checkpoint is evaluated on all three tasks. We report per-task success rates and their average, with the full evaluation protocol and success criteria provided in Appendix~\ref{app:realworld_setup}.

\begin{table}[t]
\centering
\small
\caption{Real-world manipulation performance under different pretraining schemes. Success rates are reported in percent. Each task is evaluated with $20$ trials per policy.}
\label{tab:realworld_performance}
\begin{tabular}{llccccc}
\toprule
Model & Pretrain & Press Button & Place Book & Place Drink & Avg. & Gain \\
\midrule
LA4VLA-1B & No & 60.0 & 15.0 & 40.0 & 38.3 & +0.0 \\
LA4VLA-1B & VLA & 50.0 & 40.0 & 55.0 & 48.3 & +10.0 \\
LA4VLA-1B & LA & \textbf{85.0} & \underline{65.0} & \textbf{95.0} & \underline{81.7} & \underline{+43.4} \\
LA4VLA-1B & MixPT & \underline{75.0} & \textbf{85.0} & \underline{90.0} & \textbf{83.3} & \textbf{+45.0} \\
\bottomrule
\end{tabular}
\end{table}

Table~\ref{tab:realworld_performance} shows that LA pretraining strengthens real-world manipulation ability. Averaged over the three tasks, LA4VLA-1B achieves 38.3\% success without pretraining, while VLA pretraining improves performance to 48.3\%. LA pretraining yields a much higher average success of 81.7\%, with consistent gains across all three tasks. Combining LA and VLA pretraining via MixPT further improves performance to 83.3\%. These results show that the benefits of language-action pretraining extend to real-world manipulation, substantially improving success rates over both no-pretraining and VLA-pretraining baselines, with further gains when combined with VLA pretraining through MixPT.

\subsection{Robustness to Visual Perturbations}
\label{sec:exp_generalization}

The real-world evaluation above uses visual observations without artificial perturbations. We next test whether LA pretraining also helps when the visual input is degraded at test time. We reuse the same four finetuned real-world policies from the previous experiment and evaluate them with Gaussian noise added to the input images. These policies correspond to direct downstream finetuning without pretraining, VLA pretraining, LA pretraining, and MixPT. The evaluation is conducted on the \textit{Press Button} and \textit{Place Drink} tasks, with $20$ trials per task and policy. We use Gaussian noise with standard deviation $5$ for \textit{Press Button} and $15$ for \textit{Place Drink}, setting the noise level separately because the two tasks differ in their sensitivity to visual perturbations.

\begin{table}[t]
\centering
\small
\caption{Robustness of real-world policies under Gaussian image noise. Success rates are reported in percent. Each task is evaluated with $20$ trials per policy.}
\label{tab:realworld_generalization}
\begin{tabular}{llcccc}
\toprule
Model & Pretrain & Press Button & Place Drink & Avg. & Gain \\
\midrule
LA4VLA-1B & No & 40.0 & 15.0 & 27.5 & +0.0 \\
LA4VLA-1B & VLA & 20.0 & 65.0 & 42.5 & +15.0 \\
LA4VLA-1B & LA & \underline{55.0} & \textbf{80.0} & \underline{67.5} & \underline{+40.0} \\
LA4VLA-1B & MixPT & \textbf{65.0} & \underline{75.0} & \textbf{70.0} & \textbf{+42.5} \\
\bottomrule
\end{tabular}
\end{table}

Table~\ref{tab:realworld_generalization} shows that LA pretraining improves robustness under degraded visual observations. Without pretraining, LA4VLA-1B achieves 27.5\% average success, while VLA pretraining improves it to 42.5\%. LA pretraining further boosts performance to 67.5\%, and MixPT achieves the best result at 70.0\%. Compared with both no pretraining and VLA pretraining, LA and MixPT maintain substantially higher average success under visual noise. This shows that language-action pretraining remains beneficial when visual observations are degraded, especially when used alone or combined with VLA pretraining.

\subsection{Instruction-Following Behavior and Representation Analysis}
\label{sec:exp_behavior}

We further analyze whether LA pretraining changes the policy's instruction-following behavior and internal action representations. First, we apply the direction-following protocol and metrics from Section~\ref{sec:visual_cue_dependence} to test whether the predicted action direction follows the language instruction. Second, we examine whether the internal representations used for action decoding distinguish different language instructions. 

\noindent \textbf{Evaluation settings.}
The LA-pretrained policy predicts actions from the language instruction and robot state, with visual input masked. We evaluate two settings adapted from Section~\ref{sec:visual_cue_dependence}. In the \textit{LA-pretrained} condition, the policy receives the original instruction and robot state from the LA training sample. In the \textit{Conflicting-state} condition, only the instruction specifies the target direction, while the robot state is replaced by the state from a different atomic instruction-action pair whose action trajectory moves in the opposite direction. Since visual input is masked, the robot state is the only non-language input that can conflict with the instruction.

\begin{figure*}[t]
\centering
\includegraphics[width=\textwidth]{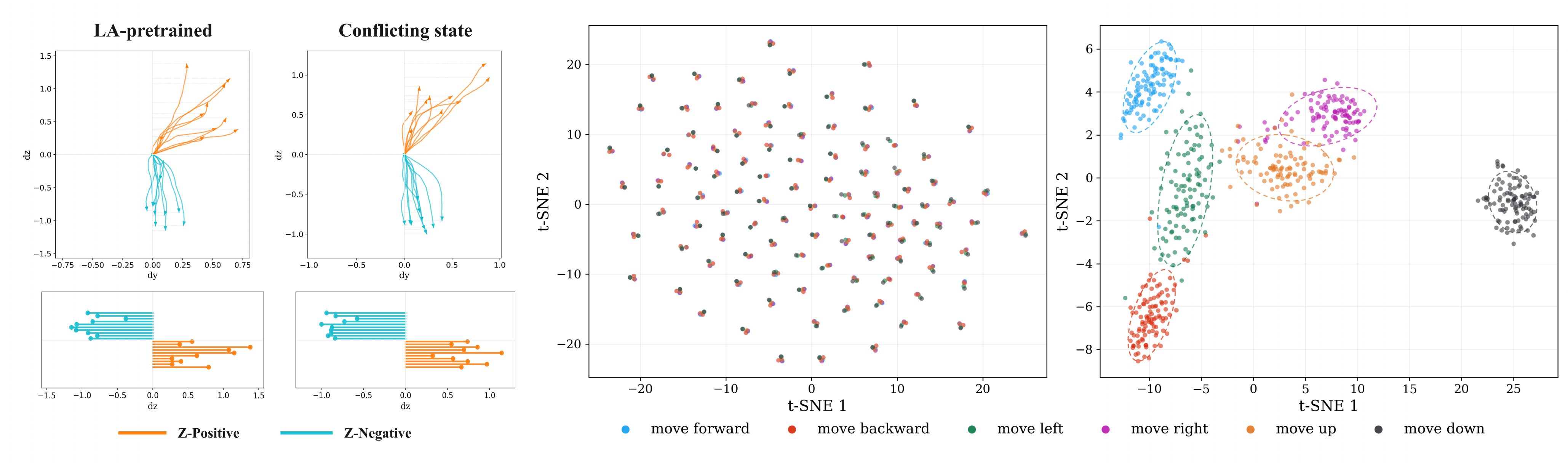}
\caption{Direction-following behavior and representation analysis of the LA-pretrained policy. \textbf{Left panel:} qualitative direction-following under the \textit{LA-pretrained} and \textit{Conflicting-state} conditions for a representative pair of opposite commands along the $z$ axis. The top row shows predicted trajectories in the $dy$--$dz$ plane, and the bottom row shows endpoint projections along the tested $z$ axis. \textbf{Right panel:} t-SNE visualization of the policy representations before action decoding, colored by direction command. The VLA-trained policy maps different commands to mixed clusters, whereas the LA-pretrained policy forms clearer command-aligned clusters.}
\label{fig:la_analysis}
\end{figure*}

\begin{table}[ht]
\centering
\caption{Quantitative direction-following results of the LA-pretrained policy under masked visual input. Metrics are computed over 100 direction-following cases for each evaluation condition.}
\label{tab:la_behavior}
\begin{tabular}{lcccc}
\toprule
Condition & DAR \(\uparrow\) & DCS \(\uparrow\) & SR \(\uparrow\) & SS \(\uparrow\) \\
\midrule
LA-pretrained & \textbf{0.99} & \textbf{0.85} & \textbf{1.46} & \textbf{0.20} \\
Conflicting state & \underline{0.95} & \underline{0.80} & \underline{1.41} & \underline{0.16} \\
\bottomrule
\end{tabular}
\end{table}

\noindent \textbf{Direction-following results.}
The left panels of Figure~\ref{fig:la_analysis} visualize the predicted trajectories and endpoint projections under the \textit{LA-pretrained} and \textit{Conflicting-state} conditions for the same representative pair of opposite commands as in Section~\ref{sec:visual_cue_dependence}. Table~\ref{tab:la_behavior} reports the corresponding directional metrics. Under the \textit{LA-pretrained} condition, the policy predicts action directions that closely follow the instruction, achieving a DCS of $0.85$ and a DAR of $0.99$. The trajectories also show clear endpoint separation between opposite instructions, indicating reliable instruction-conditioned direction following without visual input. The \textit{Conflicting-state} condition further tests whether this behavior remains when the input robot state is replaced with the state from an instruction-action pair whose motion direction is opposite to the current instruction. In this setting, DCS remains high at $0.80$, close to the \textit{LA-pretrained} value of $0.85$. 
This indicates that the predicted action direction follows the instruction rather than being dominated by the substituted robot state. 
Overall, these results show that LA pretraining yields more reliable instruction-conditioned direction following, even under the conflicting-state condition.

\noindent \textbf{Representation analysis.}
We next examine whether the direction-following results are reflected in the internal representations used for action prediction. For both policies, we extract the policy representations fed into the action head and project them with t-SNE. The representations are collected under masked visual input with the same robot states and direction instructions along the three Cartesian axes. The resulting t-SNE visualizations are shown in the right panels of Figure~\ref{fig:la_analysis}.

The two policies organize this representation space differently. For the VLA-trained policy, the projected representations form many small clusters, each containing multiple command directions. This suggests that the representations are primarily grouped by robot state, while the instruction has only a weak effect on the representation used for action decoding. This is consistent with its weak direction following under the Visual-removed-input reference in Table~\ref{tab:la_behavior}. In contrast, the LA-pretrained policy forms clearer clusters aligned with the commanded directions, and opposite commands fall into distinct regions. This suggests that LA pretraining induces a more instruction-conditioned action representation, consistent with the direction-following results above.

Taken together, the direction-following and representation results indicate that LA pretraining produces a policy whose predicted actions are more strongly conditioned on the language instruction. The policy predicts instruction-aligned directions without visual input, maintains this behavior when the input robot state is replaced with state information from an opposite-direction instruction-action pair, and forms action-decoding representations that are more clearly separated according to the instructed direction. These properties are obtained before downstream finetuning and provide behavioral and representational support for the downstream gains of policies with LA pretraining.

\section{Conclusion}

We presented LA4VLA, a language-action pretraining framework for building robust Vision-Language-Action policies. Standard VLA pretraining jointly maps visual observations and language instructions to actions, which entangles visual grounding with action learning and dilute the supervision for learning how language constrains actions. LA4VLA addresses this issue by constructing vision-agnostic atomic instruction-action pairs from existing robot demonstrations and using them for language-action pretraining. This process yields LA-33K, a 33K-episode Language-Action dataset derived without additional robot data collection. By removing visual observations during LA pretraining while preserving action supervision, the policy learns how language constrains action execution and acquires reusable language-conditioned action priors that support downstream VLA learning. Across simulation benchmarks and real-world manipulation tasks, LA pretraining consistently strengthens downstream VLA policies, outperforms VLA-only pretraining under comparable settings, and provides additional gains when combined with standard VLA training. The benefits transfer across VLA architectures and are especially evident under visual perturbations. 
Overall, our findings highlight the importance of making language-action supervision explicit rather than leaving it entangled within joint vision-language-action training.

\clearpage

\bibliographystyle{unsrt}
\bibliography{main}

\clearpage

\beginappendix

\section{LA Dataset Construction Details}
\label{app:dataset_construction}

This appendix complements the dataset construction pipeline described in Section~\ref{sec:la_dataset}. The main text presents the overall flow from original VLA data to the final LA dataset. Here we provide the implementation details for keyframe detection, atomic-action design criteria, prompt construction, VLM-based temporal segmentation, primitive definitions, and human verification.

\subsection{Keyframe Detection}

\noindent \textbf{Purpose.}
Keyframes provide temporal cues that can assist the segmentation model in identifying candidate atomic-action phases. We use them as auxiliary hints in the segmentation prompt: gripper transition frames often correspond to grasp or release/place phases, while static frames indicate near-stationary intervals that commonly occur around alignment, grasping, placing, or phase transitions.

\noindent \textbf{Static keyframes.}
Static keyframes mark short intervals in which the arm-state variation remains low. Given an episode state trajectory $\{\mathbf{s}_t\}_{t=1}^{T}$, where $\mathbf{s}_t$ denotes the robot state at frame $t$, we remove the scalar gripper channel and compute adjacent-frame arm-state differences,
\[
    \Delta \mathbf{s}_t = \mathbf{s}_{t+1}^{\mathrm{arm}} - \mathbf{s}_t^{\mathrm{arm}}.
\]
Subsequently, the obtained sequences are smoothed via Savitzky-Golay filter and normalized, and then aggregated into a scalar motion magnitude $m_t$ via mean absolute value. A frame is treated as a static candidate when $m_t$ falls below an adaptive threshold $\theta_s = r_s \cdot \mathrm{median}(m)$. For window-based detection, we also require that a sufficient proportion of frames within the local window satisfy this low-motion condition. Consecutive static candidates are merged, and the midpoint of each merged interval is used as the static keyframe. In our default setting, we use a Savitzky--Golay filter with window length $7$ and polynomial order $3$, symmetric min-max normalization, weighted-mean aggregation, $r_s=0.5$, window size $5$, and $r_w=0.8$.

\noindent \textbf{Gripper transition keyframes.}
Gripper transition keyframes mark frames where the gripper changes between open and closed states. We detect these transitions from the scalar gripper channel $g_t$. After min-max normalization, we scan the trajectory with a local temporal window. For each candidate frame $t$, we compare the mean gripper values before and after the frame, denoted as $\bar{g}^{-}_t$ and $\bar{g}^{+}_t$, respectively. A frame is selected as a transition keyframe if the two window means straddle the gripper threshold $\theta_g$ — i.e., $\bar{g}^{-}_t < \theta_g < \bar{g}^{+}_t$ or $\bar{g}^{+}_t < \theta_g < \bar{g}^{-}_t$. The resulting event is then labeled as either gripper-open or gripper-close, following the dataset's specific convention for gripper states (e.g., whether 0 indicates open or closed). In our default setting, we use $\theta_g=0.8$ and a window size of $n=3$ frames on each side. Transition events closer than $2n$ frames are filtered to reduce duplicates caused by noise.

\noindent \textbf{Algorithm.}
The resulting detector combines low-motion intervals and gripper-state transitions:

\begin{algorithm}[ht]
\caption{Keyframe detection from robot state trajectories}
\label{alg:keyframe_detection}
\begin{algorithmic}[1]
\Require Robot state trajectory $\mathbf{s}_{1:T}$ and timestamps $\tau_{1:T}$
\Ensure Sorted keyframe list $\mathcal{K}$
\State Split $\mathbf{s}_{1:T}$ into arm states $\mathbf{s}^{\mathrm{arm}}_{1:T}$ and gripper states $g_{1:T}$
\State Compute arm-state differences $\Delta \mathbf{s}_t = \mathbf{s}^{\mathrm{arm}}_{t+1} - \mathbf{s}^{\mathrm{arm}}_t$
\State Smooth $\Delta \mathbf{s}_t$ via Savitzky-Golay filter and normalize it
\State Aggregate $\Delta \mathbf{s}_t$ into scalar motion magnitude $m_t$
\State Select static candidates whose local-window motion magnitude remains below $\theta_s$
\State Merge consecutive static candidates and keep the midpoint of each merged interval
\State Normalize $g_{1:T}$ and compare before/after window means around each candidate frame
\State Select gripper-transition candidates whose window means straddle the gripper threshold $\theta_g$
\State Remove duplicate gripper-transition candidates within $2n$ frames
\State Convert retained frame indices to timestamps and sort them into $\mathcal{K}$
\State \Return $\mathcal{K}$
\end{algorithmic}
\end{algorithm}

\noindent \textbf{Prompt formatting.}
Detected keyframes are inserted into the VLM prompt as timestamped state events, for example:
\begin{quote}
\small\ttfamily
- 1.20s: static \\
- 1.80s: gripper\_close \\
- 3.40s: gripper\_open
\end{quote}
These events are provided as temporal hints for segmentation rather than hard boundaries, allowing the VLM to jointly use robot-state cues and visual evidence when determining segment boundaries.

\subsection{Atomic-Action Design Criteria}
\label{app:taxonomy_design}

\noindent \textbf{Design criteria.}
Table~\ref{tab:atomic_taxonomy} in the main text summarizes the high-level atomic-action taxonomy. The taxonomy is designed to define short segments that support language-action supervision and is guided by three criteria:
\begin{itemize}
    \item \textbf{Local temporal boundary.} Each segment should cover a short contiguous time interval and avoid spanning multiple task phases.
    \item \textbf{Observable physical change.} Each primitive should correspond to a meaningful change in robot-object interaction, object pose, gripper status, or end-effector pose.
    \item \textbf{Trajectory-grounded instruction.} Each instruction should correspond to a motion pattern that can be inferred from the observed trajectory.
\end{itemize}

These criteria define the atomic-action vocabulary and guide both VLM-based segmentation and human verification.

\subsection{Prompt Construction Details}

\noindent \textbf{Prompt components.}
The prompt guides the VLM to identify short atomic segments with structured labels and vision-agnostic instructions. It includes the original task description, sampled multi-view video frames, keyframe events, the atomic-action vocabulary, primitive definitions, category assignment rules, and the required JSON output schema. The prompt also instructs the VLM to jointly reason over all available views.

\noindent \textbf{Design goals.}
The prompt serves three roles in dataset construction. First, it reduces label drift by constraining outputs to a predefined atomic-action taxonomy. Second, it demonstrates the division of atomic actions at different granularities, including relatively short gripper actions, relatively long transport or reorient actions, and composite actions. Third, it requires both a structured action label and a natural-language instruction. The instruction is written in a vision-agnostic form, describing action type, direction, and optional object or object-status information, while avoiding object identity, appearance, material, and environmental details. This format supports LA pretraining where visual input is masked.

\noindent \textbf{Keyframe constraints.}
Keyframes provide temporal cues for identifying atomic-action boundaries. Gripper open/close events indicate likely manipulation transitions, while static keyframes often correspond to alignment, grasping, placing, or phase transitions. These cues are used as soft hints together with visual observations when determining segment boundaries.

\noindent \textbf{Prompt template.}
The following abbreviated template summarizes the segmentation input structure:
\begin{quote}
\small
\begin{tabular}{p{0.95\linewidth}}
\toprule
\textbf{Task context:} original long-horizon task description and sampled multi-view video frames. \\
\midrule
\textbf{Action vocabulary:} grasp, place, lift, lower, transport, wrist\_rotate, reorient, push, pull, press, move, rotate, and wait. \\
\textbf{Object-status tags:} HOLDING and EMPTY, depending on whether an object is held by the gripper. \\
\textbf{Primitive definitions:} specific definitions and descriptions of different action primitives \\
\textbf{Keyframe cues:} timestamped static, gripper-open, and gripper-close events. \\
\textbf{Instruction rules:} describe action type, direction, and optional object-status information while avoiding object identity and appearance details. \\
\textbf{Output fields:} \texttt{subaction}, \texttt{instruction}, \texttt{start}, and \texttt{end} in JSON format. \\
\textbf{Granularity examples:} decomposition examples such as decompose place cup into move $\rightarrow$ grasp $\rightarrow$ lift $\rightarrow$ transport $\rightarrow$ place. \\
\bottomrule
\end{tabular}
\end{quote}

\subsection{VLM-based Temporal Segmentation}

\noindent \textbf{Model and inputs.}
We use Qwen-3-VL-Plus to propose temporal segments for each long-horizon demonstration. Video frames are sampled at $2$ FPS and resized to $224 \times 224$. When multiple camera views are available, each view is provided as a separate video stream with an associated textual view label. The VLM is provided with the original task description, the segmentation prompt, and formatted keyframe events as additional temporal cues.

\noindent \textbf{Output format.}
The VLM outputs a JSON list of candidate atomic-action segments. Each segment contains a structured primitive label, a vision-agnostic instruction, and start/end timestamps:
\begin{quote}
\small
\begin{tabular}{p{0.95\linewidth}}
\toprule
\texttt{[} \\
\quad \texttt{\{} \\
\quad\quad \texttt{"subaction": "lift",} \\
\quad\quad \texttt{"instruction": "lift the object upward while holding it",} \\
\quad\quad \texttt{"start": 3.50,} \\
\quad\quad \texttt{"end": 5.00} \\
\quad \texttt{\}} \\
\texttt{]} \\
\bottomrule
\end{tabular}
\end{quote}
We use structured JSON outputs with a sampling temperature of $0.2$ to improve consistency. Failed or malformed outputs are automatically retried.

\noindent \textbf{Proposal generation.}
The VLM is used as a proposal generator for candidate temporal segmentation. It integrates visual observations, task context, keyframe cues, and taxonomy constraints to produce initial segment proposals. These proposals are subsequently filtered and refined through human verification and quality control before being included in the final LA dataset.

\noindent \textbf{Cost Analysis.} 
Segmenting 1/10 of the DROID dataset (9,560 episodes) with Qwen-3-VL-Plus incurred a token consumption of 44.29M input tokens and 3.385M output tokens. On average, each episode required 4.633K input tokens and 0.354K output tokens, demonstrating the cost feasibility of using a VLM for dataset segmentation.

\subsection{Primitive Definitions}

\begin{figure}[ht]
\centering
\includegraphics[width=0.88\linewidth]{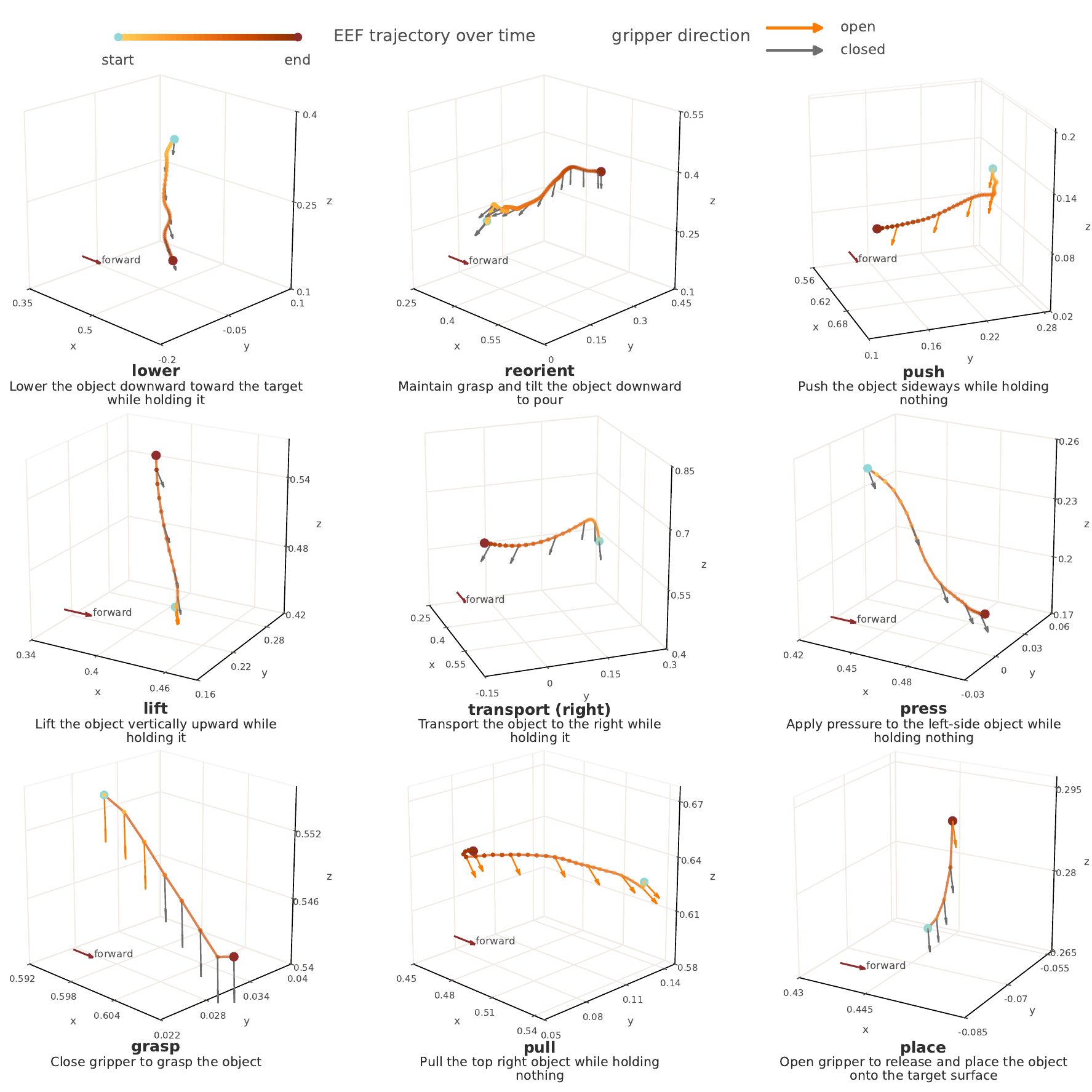}
\caption{
The visualization of more action primitives
}
\label{fig:atomic_action}
\end{figure}

The specific definitions of the action primitives are provided in Table \ref{tab:primitive_definitions}, and visualizations of more atomic actions are provided in Figure \ref{fig:atomic_action}. Further details are elaborated below.

\noindent \textbf{Definition convention.}
We characterize each primitive by its dominant physical effect within an atomic segment, using end-effector motion, gripper state, object status, and contact mode as cues. For directional translations, we adopt a robot base-frame convention: $+x$ denotes forward, $-x$ backward, $+y$ left, $-y$ right, and $+z$ upward, while $-z$ denotes downward. Under this convention, primitives such as \textit{transport} and \textit{move} are annotated with respect to base-frame directions.

\noindent \textbf{Dominant primitive rule.}
Real robot demonstrations may contain overlapping motion patterns. For example, a segment may involve both transport and minor reorientation, or short alignment motions before grasping. In such cases, we assign a single dominant primitive label based on the primary physical effect, while the natural-language instruction may additionally describe secondary effects when relevant. For instance, a segment with primarily horizontal displacement is labeled as \textit{transport} and may be described as "transport the object left while keeping it upright." Similarly, a grasp segment may include a short alignment phase and be described as "align with and grasp the object."

\noindent \textbf{Rotational primitives.}
Rotational actions are represented either as axis-aligned rotations or target-oriented reorientation, depending on which description better reflects the trajectory. Wrist-axis rotations applied to a held object are denoted as \textit{wrist\_rotate}, while orientation-changing actions are denoted as \textit{reorient}, e.g., upright, flat, vertical, or horizontal.

\begin{table}[ht]
\centering
\caption{Primitive definitions and typical motion patterns in the LA dataset.}
\label{tab:primitive_definitions}
\begin{tabular}{p{0.18\linewidth}p{0.74\linewidth}}
\toprule
Primitive & Definition and Typical Motion Pattern \\
\midrule
grasp & Align with a target object and close the gripper to establish a grasp. \\
place / release & Move a held object to a target pose or support region and open the gripper. \\
lift & Move a held object primarily along $+z$. \\
lower & Move a held object primarily along $-z$. \\
transport & Move a held object in the horizontal plane along base-frame directions. \\
wrist\_rotate & Rotate a held object around the tool axis (clockwise or counterclockwise). \\
reorient & Adjust a held object's orientation toward a target state (e.g., upright, flat). \\
push & Apply contact-based motion to move an object away from the end effector. \\
pull & Apply contact-based motion to move an object toward the end effector. \\
press & Apply downward contact force to activate or depress a target. \\
move & Move the empty gripper in free space using base-frame directions. \\
rotate\_free & Rotate the empty gripper in free space around its axis. \\
reorient\_free & Adjust the empty gripper orientation toward a target state. \\
\bottomrule
\end{tabular}
\end{table}

\subsection{Human Verification Protocol}

\noindent \textbf{Annotation target.}
Human verification is applied to the VLM-generated candidate atomic-action segments before they are included in the LA dataset. For each candidate, annotators inspect the corresponding video segment together with the structured subaction label and the generated instruction. They evaluate whether the temporal boundary, action label, and instruction are mutually consistent. This step filters noisy VLM proposals, including over-segmented clips, merged primitives, imprecise boundaries, and instructions that are not well aligned with the observed action.

\noindent \textbf{Quality score.}
Each segment receives a quality score from $0$ to $3$, as shown in Table~\ref{tab:quality_score}. The score measures whether the segment is suitable for LA dataset training in terms of temporal boundary quality, action-label correctness, and language-action alignment.

\begin{table}[ht]
\centering
\caption{Quality score used for human verification.}
\label{tab:quality_score}
\begin{tabular}{clp{0.68\linewidth}}
\toprule
Score & Meaning & Interpretation \\
\midrule
0 & Unusable & The segment is severely misaligned or does not contain a meaningful atomic action. \\
1 & Low quality & The segment contains relevant motion but the boundary, label, or instruction is too noisy for reliable training. \\
2 & Acceptable & The segment is mostly correct and useful for training with minor imperfections. \\
3 & High quality & The segment, subaction label, and instruction are well aligned. \\
\bottomrule
\end{tabular}
\end{table}

\noindent \textbf{Confidence score.}
Annotators also provide a confidence score from $0$ to $2$, as shown in Table~\ref{tab:confidence_score}. This score records annotator certainty in the quality judgment and is used to monitor ambiguous cases and annotation reliability. It is not used in the final filtering stage. The average confidence score is $1.97$, indicating generally high annotator certainty.

\begin{table}[ht]
\centering
\caption{Confidence score recorded during human verification.}
\label{tab:confidence_score}
\begin{tabular}{clp{0.68\linewidth}}
\toprule
Score & Level & Interpretation \\
\midrule
0 & Low & The segment is ambiguous or difficult to assess. \\
1 & Medium & The annotator can assess the segment but some uncertainty remains. \\
2 & High & The annotator is confident in the quality judgment. \\
\bottomrule
\end{tabular}
\end{table}

\noindent \textbf{Agreement evaluation.}
To evaluate annotation consistency, $10\%$ of the generated segments are annotated by at least two annotators. We measure inter-annotator agreement using Krippendorff's alpha on an interval scale and obtain $\alpha = 0.7794$. This result suggests a reasonably consistent level of agreement across annotators for the verification task.

\noindent \textbf{Filtering rule.}
The cleaned LA dataset retains segments with quality score at least $2$ and discards segments with quality score below $2$. For samples annotated by multiple annotators, we use the average quality score for filtering. Thus, retained segments must be rated at least acceptable in terms of temporal boundary quality, action-label correctness, and language-action alignment. This rule removes low-quality VLM proposals while preserving segments suitable for LA dataset training.

\section{Real-World Experimental Setup}
\label{app:realworld_setup}

\noindent \textbf{Platform.}
The real-world experiments use a UFactory xArm6 arm with a parallel-jaw gripper. The policy receives two RGB camera views, one from a wrist-mounted camera and one from a third-person camera.

\noindent \textbf{Tasks.}
We design three language-conditioned manipulation tasks. In \textit{Press Button}, four buttons are placed at different positions on the table, and the policy must press the button specified by the instruction. In \textit{Place Book}, the policy must insert a book into one of three shelf positions specified by the instruction. In \textit{Place Drink}, the policy must place a drink into the cell of a $3\times3$ grid specified by the instruction. Within each task, the initial scene and observation are shared across target goals, so the visual input cannot disambiguate the target and the instruction alone determines which goal to execute.

\noindent \textbf{Demonstrations and training.}
We collect $100$ teleoperated demonstrations per task, $300$ in total. The two compared policies are finetuned on the same demonstrations under the same protocol, and each policy uses a single multi-task checkpoint that is evaluated on all three tasks.

\noindent \textbf{Evaluation protocol.}
In \textit{Press Button}, the four target buttons are evaluated in a fixed cyclic order: top-right, bottom-right, bottom-left, and top-left, repeated for $5$ rounds, giving $20$ trials in total. In \textit{Place Book}, the three shelf positions are tested $7$, $6$, and $7$ times, giving $20$ trials. In \textit{Place Drink}, four target cells are each evaluated $5$ times, giving $20$ trials. Each trial is capped by a step limit of $800$ low-level control steps for Press Button, $1{,}200$ for Place Book, and $1{,}000$ for Place Drink; for example, with an action horizon of $25$, $800$ steps correspond to approximately $32$ action chunks.

\noindent \textbf{Success criteria.}
Press Button counts as a success if the commanded button is pressed at least once within the step limit, and as a failure otherwise. Place Book counts as a success if the book is inserted to at least half of the shelf depth. Place Drink counts as a success if the drink is placed into the commanded cell in an upright and stable configuration without tipping over within the step limit. Prematurely opening the gripper, causing the drink to spill, is also considered a failure.

\section{Metrics for Instruction Following Analysis}
\label{app:metrics_for_instruction_analysis}

\noindent \textbf{Notations.}
The evaluation set is denoted as
\[
\mathcal{D}=\{(y_i,\hat{\mathbf{p}}_i,\mathbf{g}_i,\mathbf{u}_i)\}_{i=1}^{N},
\]
where $y_i \in \mathcal{Y}$ is the instruction direction label,
$\hat{\mathbf{p}}_i \in \mathbb{R}^{3}$ is the predicted endpoint displacement,
$\mathbf{g}_i \in \mathbb{R}^{3}$ is the ground-truth endpoint displacement under the corresponding instruction, and
$\mathbf{u}_i \in \mathbb{R}^{3}$ is the unit vector corresponding to the instruction-specified canonical direction.
For example, $\mathbf{u}_i=(1,0,0)$ for $x^+$ and $\mathbf{u}_i=(-1,0,0)$ for $x^-$.
The endpoint displacements are obtained by summing translational increments over the prediction horizon:
\[
\hat{\mathbf{p}}_i = \sum_{t=1}^{T} \Delta \hat{\mathbf{x}}_{i,t},
\qquad
\mathbf{g}_i = \sum_{t=1}^{T} \Delta \mathbf{x}^{\mathrm{GT}}_{i,t}.
\]

\noindent \textbf{Directional Alignment Rate (DAR).}
DAR measures the fraction of trials whose predicted endpoint displacement lies in the half-space aligned with the canonical instruction direction:
\[
\mathrm{DAR}
=
\frac{1}{N}
\sum_{i=1}^{N}
\mathbb{I}
\left[
\hat{\mathbf{p}}_i^\top \mathbf{u}_i > 0
\right].
\]

\noindent \textbf{Direction Consistency Score (DCS).}
DCS measures the cosine similarity between the predicted endpoint displacement and the ground-truth endpoint displacement:
\[
\mathrm{DCS}
=
\frac{1}{N}
\sum_{i=1}^{N}
\frac{
\hat{\mathbf{p}}_i^\top \mathbf{g}_i
}{
\|\hat{\mathbf{p}}_i\|_2 \, \|\mathbf{g}_i\|_2
}.
\]
Since $\mathbf{g}_i$ is collected under the corresponding instruction, DCS evaluates whether the predicted motion follows the instruction-conditioned demonstrated direction. In implementation, trials with near-zero predicted or ground-truth displacement norm are assigned a cosine value of $0$.

\noindent \textbf{Separability Ratio (SR).}
SR compares two quantities: the average distance between predictions with different instruction labels, and the average distance between predictions with the same instruction label. For each instruction label $k \in \mathcal{Y}$, let
\[
\mathcal{I}_k = \{ i : y_i = k \}
\]
be the indices of trials labeled as direction $k$.

First compute the average within-direction distance by averaging pairwise distances within each direction label and then averaging over direction labels:
\[
d_{\mathrm{intra}}
=
\frac{1}{|\mathcal{Y}|}
\sum_{k \in \mathcal{Y}}
\frac{2}{|\mathcal{I}_k|(|\mathcal{I}_k|-1)}
\sum_{\substack{i,j \in \mathcal{I}_k \\ i<j}}
\|\hat{\mathbf{p}}_i-\hat{\mathbf{p}}_j\|_2.
\]
Then compute the average between-direction distance by averaging pairwise distances for each direction pair and then averaging over direction pairs:
\[
d_{\mathrm{inter}}
=
\frac{2}{|\mathcal{Y}|(|\mathcal{Y}|-1)}
\sum_{\substack{a,b\in\mathcal{Y}\\a<b}}
\frac{1}{|\mathcal{I}_a||\mathcal{I}_b|}
\sum_{i\in \mathcal{I}_a}
\sum_{j\in \mathcal{I}_b}
\|\hat{\mathbf{p}}_i-\hat{\mathbf{p}}_j\|_2.
\]
Here $a<b$ simply means that each pair of different direction labels is counted once. The separability ratio is then defined as follows, and here $\epsilon$ denotes a small positive constant used for numerical stability:
\[
\mathrm{SR}
=
\frac{
d_{\mathrm{inter}}
}{
d_{\mathrm{intra}}+\epsilon
}.
\]
A higher ratio means that the predicted endpoint displacements are more separable by instruction direction.

\noindent \textbf{Silhouette Score (SS).}
SS measures how well endpoint displacements cluster according to instruction direction.
For each trial $i$, define the average distance to trails with the same instruction direction as
\[
a_i
=
\frac{1}{|\mathcal{I}_{y_i}|-1}
\sum_{\substack{j\in \mathcal{I}_{y_i}\\j\neq i}}
\|\hat{\mathbf{p}}_i-\hat{\mathbf{p}}_j\|_2,
\]
and the nearest average distance to another instruction-direction cluster as
\[
b_i
=
\min_{k\neq y_i}
\frac{1}{|\mathcal{I}_{k}|}
\sum_{j\in \mathcal{I}_{k}}
\|\hat{\mathbf{p}}_i-\hat{\mathbf{p}}_j\|_2.
\]
The silhouette value for trial $i$ is
\[
s_i
=
\frac{
b_i-a_i
}{
\max(a_i,b_i)+\epsilon
},
\]
and the overall silhouette score is
\[
\mathrm{SS}
=
\frac{1}{N}
\sum_{i=1}^{N}
s_i.
\]
Higher SS indicates that endpoint displacements are closer to trials from the same instruction direction than to trials from other directions.

\end{document}